\documentclass[10pt,twocolumn,letterpaper]{article}

\usepackage{cvpr}
\usepackage{times}
\usepackage{epsfig}
\usepackage{graphicx}
\usepackage{amsmath}
\usepackage{amssymb}
\usepackage{slashbox}
\usepackage{cite}

\usepackage{multirow}
\usepackage{subfiles}
\usepackage[dvipsnames]{xcolor}
\usepackage{booktabs}
\usepackage{colortbl}

\usepackage[pagebackref=true,breaklinks=true,letterpaper=true,colorlinks,bookmarks=false]{hyperref}

\cvprfinalcopy 

\def\cvprPaperID{9330} 

\def\eqnvspace{{\vspace{-1mm}}}
\def\figvspace{{\vspace{-2mm}}}
\def\secvspace{{\vspace{-2mm}}}

\ifcvprfinal\fi
\begin{document}

\title{Noise Modeling, Synthesis and Classification for Generic Object Anti-Spoofing}

\author{Joel Stehouwer, Amin Jourabloo, Yaojie Liu, Xiaoming Liu \\
Department of Computer Science and Engineering \\
Michigan State University, East Lansing MI 48824\\
{\tt \{stehouw7, liuyaoj1, jourablo, liuxm\}@msu.edu}
}

\maketitle

\begin{abstract}
Using printed photograph and replaying videos of biometric modalities, such as iris, fingerprint and face, are common attacks to fool the recognition systems for granting access as the genuine user.
With the growing online person-to-person shopping (e.g., Ebay and Craigslist), such attacks also threaten those services, where the online photo illustration might not be captured from real items but from paper or digital screen.
Thus, the study of anti-spoofing should be extended from modality-specific solutions to generic-object-based ones.
In this work, we define and tackle the problem of Generic Object Anti-Spoofing (GOAS) for the first time.
One significant cue to detect these attacks is the noise patterns introduced by the capture sensors and spoof mediums. 
Different sensor/medium combinations can result in diverse noise patterns.
We propose a GAN-based architecture to synthesize and identify the noise patterns from seen and unseen medium/sensor combinations. 
We show that the procedure of synthesis and identification are mutually beneficial.
We further demonstrate the learned GOAS models can directly contribute to modality-specific anti-spoofing without domain transfer.
The code and GOSet dataset are available at \url{cvlab.cse.msu.edu/project-goas.html}.

\end{abstract}
\section{Introduction}
\secvspace

Anti-spoofing (\emph{i.e.}, spoof detection) is a long-standing topic in the biometrics field that empowers recognition systems to detect samples from spoofing mediums, \emph{e.g.}, printed paper or digital screen~\cite{atoum,boulkenafet2015face,boulkenafet-generalization,liu-siw}.
A similar concern may appear in online commerce websites, \emph{e.g.}, Ebay, Craigslist, which provide services to enable direct user-to-user buying and selling.
For instance, when purchasing, a customer may wonder, ``Is that a picture of a real item he owns?"
This scenario motivates a {\it broader} problem of anti-spoofing: 
\vspace{-2mm}
\begin{quote}
{\it Given an image of a generic object, such as a cup or a desk, can we automatically classify if this was captured from the real object, or through a medium, such as digital screen or printed paper?}
\end{quote}
\vspace{-2mm}

\begin{figure}[t]
\begin{center}
\includegraphics[width=\linewidth]{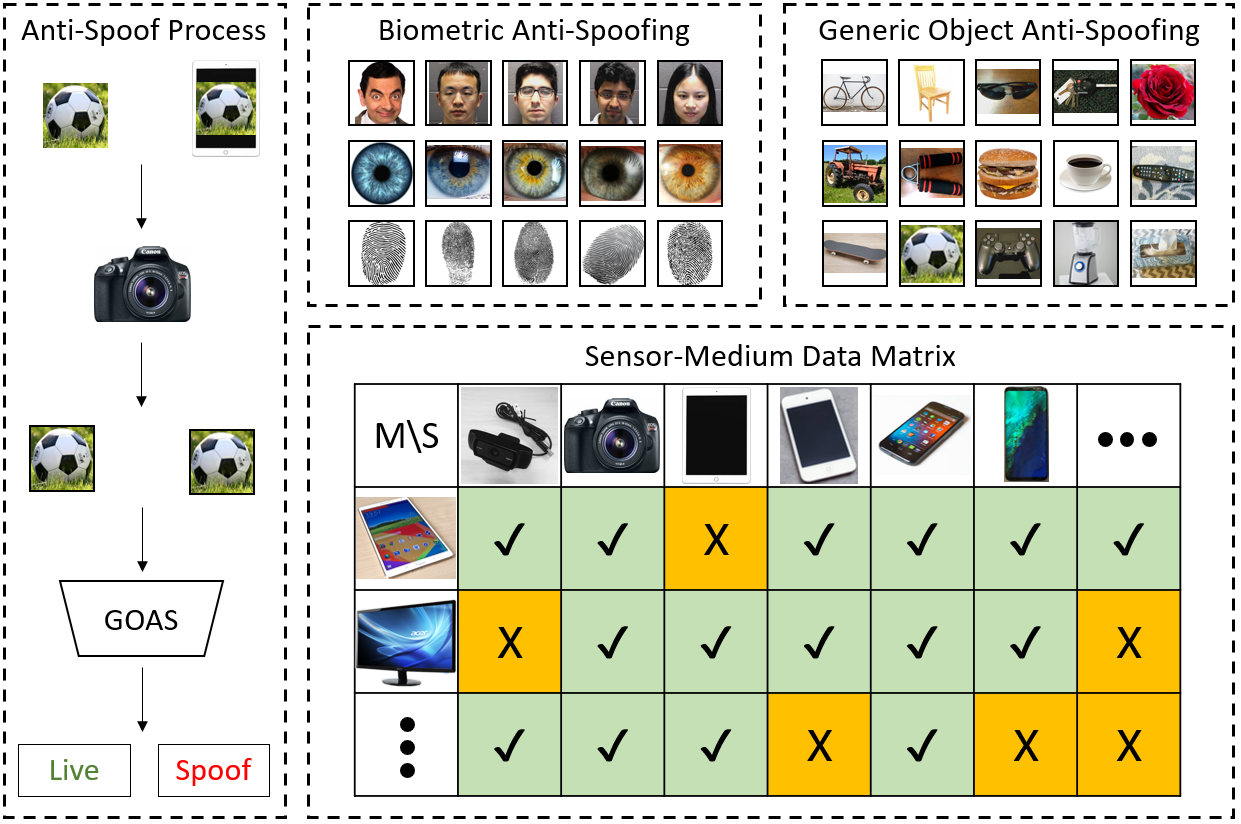}
\end{center}
\vspace{-2mm}
\caption{\small  Similarly to biometric anti-spoofing, GOAS determines if an image of an object is captured from the real object or through spoof mediums. Anti-spoofing algorithms can be sensitive to device-specific noises. Given the challenge of capturing spoof data with full combinations of sensors/mediums, we synthesize spoof images at any combination (marked as \textbf{X}), which benefits GOAS.}
\label{fig:go_motivation}
\figvspace
\vspace{-2mm}
\end{figure}

We define this problem as Generic Object Anti-Spoofing (GOAS).
With the wider variety of objects, there are richer appearance variations and greater challenges in GOAS, as shown in Fig.~\ref{fig:go_motivation}, compared to individual biometric modalities.
Successful solutions~\cite{atoum,boulkenafet-generalization,jourabloo,liu-siw,Liu_2019_CVPR,pan-eyeblink,patel-face-unlock,Perez-Cabo_2019} for modality-specific anti-spoofing are likely ineffective for GOAS.
We find that capture sensors and spoofing mediums bring certain texture patterns (\emph{e.g.}, Moir\'e pattern~\cite{patel-moire-pattern}) to all captured images, regardless of the content.
These patterns are often low-energy and regarded as ``{\it noise}''.
However, they are ubiquitous and consistent, since they result from the physical properties of the sensors/mediums and environmental conditions, such as light reflection.
We believe a proper modeling of such noise patterns will lead to effective solutions for GOAS and may contribute to modality-specific anti-spoofing tasks.
In this work, we study the fundamental low-level vision problem of {\it modeling, synthesizing, and classifying the noise patterns for tackling GOAS}.

Modeling noise patterns is a promising, yet challenging, approach for GOAS. 
In~\cite{chen-ensemble,thai-cmi,chen2020}, the camera model identification problem is studied for the purpose of digital forensics.
The properties of different capture sensors are examined thanks to the assistance of databases, such as PRNU-PAR Dataset~\cite{guera-counter-forensic} and Dresden Image Database~\cite{gloe2010dresden}.
Related topics such as noise pattern removal~\cite{abdelhamed2018high} and noise pattern modeling for face modality~\cite{jourabloo} are also investigated.
The authors of~\cite{Yang_2019_CVPR} show that simple synthesis methods for data augmentation are beneficial for the anti-spoofing task.
These prior works provide a solid base to begin the study of GOAS.
Meanwhile, we still face three major challenges:

\textbf{Complexity of spoof noise patterns:} 
The noise patterns in GOAS are related to both sensor and medium, as well as their interaction with the environment.
First, it's hard to model the interaction mathematically.
Second, the noises are ``hidden'' under large appearance variations, thus even more untraceable.
Additionally, each physical device has a unique fingerprint, though these fingerprints are similar within the same device models as shown in \cite{miroslav-spie,filler-icip}.

\textbf{Insufficient data and lack of strong labels:} 
Unlike many other computer vision tasks, spoof data for anti-spoofing cannot be collected from the Internet.
Moreover, strong labels, \emph{e.g.}, pixel-wise correspondence between spoof images and ground truth live images, is extremely difficult to obtain.
The constant development of new sensors and spoof mediums further complicates the data collection, and increases the difficulty of learning a CNN that is robust to these small, but significant variations~\cite{oulu-competition-paper}.

\textbf{Modality dependency:}
Current anti-spoofing methods are designed for a specific modality, \emph{e.g.}, face, iris, or fingerprint.
These solutions cannot be applied to a different modality.
Thus, it is desirable to have a single anti-spoofing model applicable to multiple modalities or applications.

To address these challenges, we propose a novel Generative Adversarial Network (GAN)-based approach for GOAS, consisting of three parts: GOGen, GOLab, and GoPad.
GOGen is a generator network, which learns to convert a live image to a spoof one given a target \textit{known or unknown} sensor/medium combination. 
GOGen allows for synthesis of new images with specific combinations, which helps to remedy insufficiency and imbalance issues in training data, such as the long tail problem~\cite{feature-transfer-learning-for-face-recognition-with-under-represented-data}.
GOLab serves as a multi-class classifier to identify the type of sensor and medium as well as live vs.~spoof. 
GoPad is a binary classifier for GOAS.
The three parts in this design, including the synthesis procedure and multi-class identification, contribute to our final goal of GOAS.
To properly train such a network, three novel loss functions are proposed to model the noise pattern and supervise the training.
Furthermore, we collect the first generic object dataset (GOSet) to conduct this study. 
GOSet involves $7$ camera sensors, $7$ spoof mediums, and other image variations.

To summarize, the contributions of this work include:

$\diamond$ We identify and define the new problem of GOAS.

$\diamond$ We propose a novel network architecture to synthesize unseen noise patterns that are shown to benefit GOAS.

$\diamond$ A generic object dataset (GOSet) is collected and contains live and spoof videos of $24$ objects.

$\diamond$ We demonstrate SOTA generalization performance when applying GOSet trained models to face anti-spoofing.
\section{Prior Work}
\secvspace

While there is no prior work on GOAS, we review relevant prior work from three perspectives.

\textbf{Modality-specific anti-spoofing:}
Early works~\cite{boulkenafet2015face, boulkenafet-generalization} perform texture analysis via hand-crafted features for anti-spoofing.
\cite{atoum} utilizes a patch-based CNN and score fusion to show that spoof noise can be detected in small image patches.
Similarly,~\cite{chugh2018fingerprint} uses minutiae to guide patch selection for fingerprint anti-spoofing.
Rather than detecting spoof noise, ~\cite{jourabloo} attempts to estimate and remove the spoof noise from images.
Cue-based methods incorporate domain knowledge into anti-spoofing, \emph{e.g.}, rPPG~\cite{3dmask-rppg, liu-siw}, eyeblinks~\cite{pan-eyeblink}, visual rhythms~\cite{bao-liveness-optical-flow, feng-motion-cues, pinto-visual-rhythm}, paired audio cues~\cite{chetty-audio-visual}, and pulse oximetry~\cite{reddy-pulse-oxiometry}.
A significant limitation is that each modality is domain specific; an algorithm developed for one modality cannot be applied to the others.
The closest approach to cross domain is~\cite{Lucena2017TransferLU}, via transfer learning to fine-tune on the face modality.
Our work improves upon these by utilizing generic objects, and therefore is forced to be content independent. 
Further, we learn a deep representation for the spoof noise of multiple spoof mediums, and show that these noises can be convolved with a live image to synthesize new spoof images.

\begin{figure*}[t]
\begin{center}
\includegraphics[width=\textwidth]{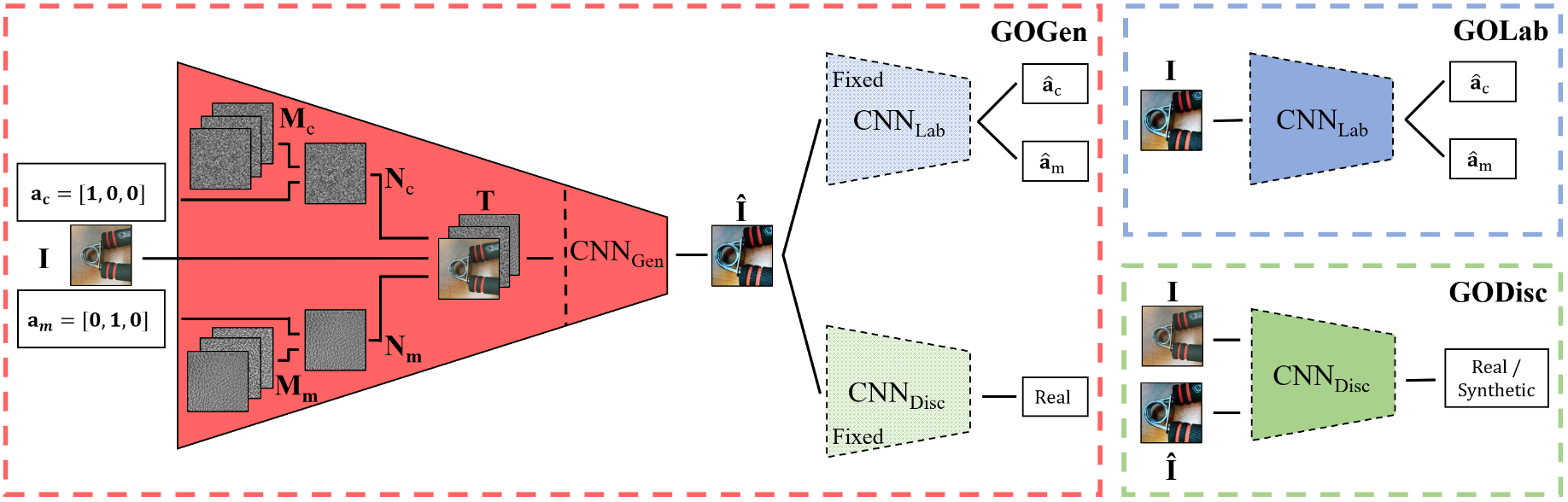}
\end{center}
\vspace{-2mm}
\caption{The overall framework of training GOGen. 
Live images are given to the generator to modify either the sensor or spoof noise. 
The resulting image is classified by the GOLab discriminator to supervise the generated images. 
An additional discriminator is used to ensure the generated images remained visually appealing and realistic.
In each section of the figure, only the solid-colored network is updated in that training step.
We alternate between training GOGen in one step and GOLab and GODisc in the next step.
Input one-hot vectors are used as a mask to select the appropriately learned noise map, which is then concatenated to the input image.}
\figvspace
\label{fig:gogen_structure}
\end{figure*}

\textbf{Noise patterns modeling:}
Modeling or extracting noise from images is challenging, since there is no canonical ground truth.
Hence some works attempt to estimate the noise via assumptions about the physical properties of the sensors and software post-processing of captured images~\cite{thai-heteroscedastic, thai-cmi}.
With these assumptions, ensemble classifiers~\cite{chen-ensemble}, hand-crafted feature based classifiers~\cite{thai-heteroscedastic, thai-cmi}, and deep learning approaches~\cite{guera-counter-forensic} are proposed to address camera model identification.
Following these, we assume that the sensor noise is image content independent.
However, we not only classify the noise in an image, but also learn a noise prototype for each sensor that can be convolved with any image to modify its ``noise footprint''.
We also address the challenge of spoof medium noise modeling and classification.
~\cite{jourabloo} estimates the spoof noise on an image, but is limited to face images and estimates the noise per image.
Hence we extend both camera model identification and spoof noise estimation works by combining both tasks within a single CNN, and by modeling a generalized representation of both the sensor and medium noises.

\textbf{Image manipulation and synthesis:}
GANs have gained increasing interest for style transfer and image synthesis tasks.
Star-GAN~\cite{star-gan} utilizes images from multiple domains and datasets to accurately manipulate images by modifying attributes. 
\cite{cycle-gan} attempts to ensure high-fidelity manipulation by requiring the generator to learn a mapping such that it recreates the original image from the synthetic one.
The work in~\cite{conditional-gan} shows that it is possible to conditionally affect the output of a GAN by feeding an extra label, {\it e.g.}, poses~\cite{disentangled-representation-learning-gan-for-pose-invariant-face-recognition}.
Here, we propose a GAN-based, targeted, content independent, image synthesis algorithm (GOGen) to alter {\it only} the high-frequency information of an image.

Similarly, image super-resolution~\cite{dong-super-res, image-super-resolution-via-deep-recursive-residual-network, memnet-a-persistent-memory-network-for-image-restoration, fsrnet-end-to-end-learning-face-super-resolution-with-facial-priors} is used to improve the visual quality and high-frequency information in an image.
\cite{lai-image-sr} uses a laplacian pyramid structure to convert a low-resolution (LR) image into a high-resolution (HR) one.
~\cite{song-image-sr} estimates an HR gradient field and uses it with an upscaled LR image to produce an HR one.
While super-resolution produces high-frequency information from low-frequency input, our GOGen aims to {\it alter} the existing high-frequency information in the input live image, which is particularly challenging given its unpredictable nature.

\section{Proposed Methods}
\secvspace

In this section, we present the details of the proposed methods, including GOGen, GODisc, and GOLab.
As shown in Fig.~\ref{fig:gogen_structure}, the overall framework adopts a GAN architecture, which is composed of a generator (GOGen) and two discriminators (GODisc and GOLab).
GOGen synthesizes additional spoof videos of any combination of sensor and medium, even \textit{unseen} combinations.
GODisc is the discriminator network to guide images from GOGen to be visually plausible. 
GOLab performs sensor and medium identification.
In addition, GOLab serves as the module to produce a final spoof detection score.
We also present GOPad, which is adapted from a traditional binary classifier used by previous anti-spoofing works, to compare with the proposed method. 
To prevent overfitting and increase the quantity of training data, the input for the networks are image patches extracted from the original images.

\subsection{GOGen: Spoof Synthesis}

In anti-spoofing, the increasing variety of sensors and spoof mediums creates a large challenge for data collection and generalization.
It is increasingly expensive to collect additional data from every combination of camera and spoof medium.
Meanwhile, the quantity, quality, and diversity of training data determine the performance and impact the generalization of deep learning algorithms.
Hence, we develop GOGen to address this need for continual data collection via synthesis of unseen combinations.

We train GOGen to synthesize new images of unseen sensor/medium combinations using knowledge learned from known combinations.
When introducing a new device, GOGen can be trained with minimal data from the new device while utilizing all previously collected data from other devices.
The generator, $\text{CNN}_{Gen}()$, converts a live image into a targeted spoof image of a specified spoof medium captured by a specified sensor. 
Specifically, the inputs of the generator are a live image $\textbf{I}\in \mathbb{R}^{H\times W}$ and two one-hot vectors specifying the sensor of the output image $\textbf{a}_c \in \mathbb{R}^{n_c}$, and the medium through which the output  would be captured $\textbf{a}_m \in \mathbb{R}^{n_m}$.
The output is a synthetic image $\hat{\textbf{I}}$.

\begin{figure}[t]
	\begin{center}
		\includegraphics[width=\linewidth]{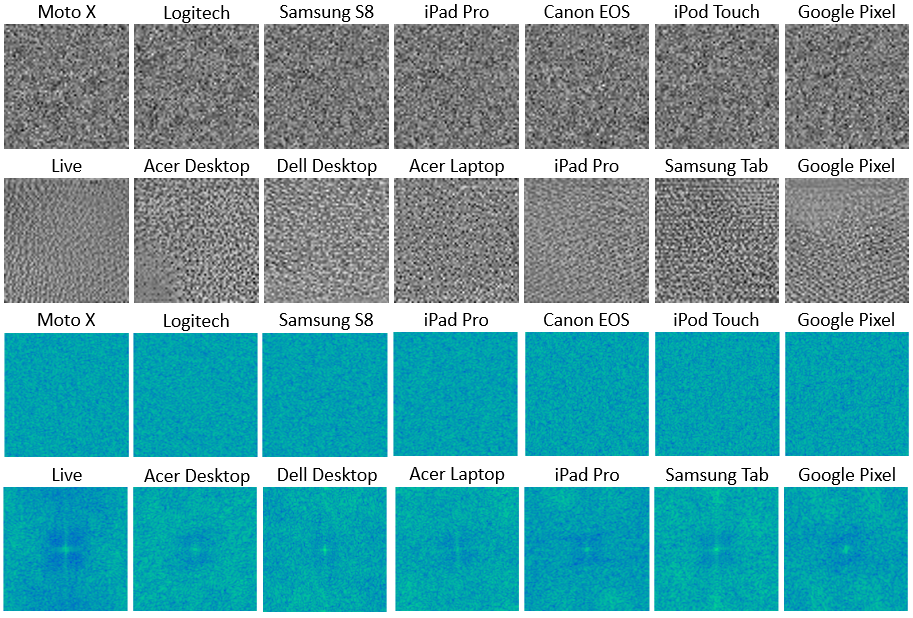}
	\end{center}
	\vspace{-2mm}
	\caption{GOGen learns noise prototypes of sensors $\textbf{M}_c$  (row $1$) and spoof mediums $\textbf{M}_m$ (row $2$). Rows $3$ and $4$ shows the $2$D FFT power spectrum of noise prototypes in rows $1$ and $2$, respectively.}
	\figvspace
	\vspace{-2mm}
	\label{fig:sensor_medium_maps}
\end{figure}

One key novelty in GOGen is the modeling of the noise from different sensors and spoof mediums.
We assume the sensor and medium noises are {\it image independent} since they are attributed to the hardware, while the noise on an image is {\it image dependent}, due to interplay between the sensor, medium, image content, and imaging environment.
To model such interplay, we denote a set of {\it image-independent} latent noise prototypes for all types of sensors $\textbf{M}_c\in \mathbb{R}^{H\times W \times n_c}$, and mediums $\textbf{M}_m\in \mathbb{R}^{H\times W \times n_m}$.
In the training, using input one-hot vectors, $\textbf{a}_c$ and $\textbf{a}_m$, we select the noise prototypes for the specific sensor-medium combination, $\textbf{N}_c,\textbf{N}_m \in \mathbb{R}^{H\times W}$, via:
\eqnvspace
\begin{equation}
\textbf{N}_c=\sum_{i=1}^{n_c}\textbf{a}_c^i\textbf{M}_c^i , \ \qquad \ \textbf{N}_m=\sum_{i=1}^{n_m}\textbf{a}_m^i\textbf{M}_m^i.\\
\end{equation}
Then, we concatenate $\textbf{I}$, $\textbf{N}_c$ and $\textbf{N}_m$ as $\textbf{T}=[\textbf{I},\textbf{N}_c,\textbf{N}_m]$ and feed $\textbf{T}$ to the generator. 
With this concatenated input, through convolution the generator mimics the interplay between the image content $\textbf{I}$, and the learnt $\textbf{N}_c$ and $\textbf{N}_m$, to generate a device-specific, image-dependent, synthetic image.
By manipulating only the sensor or the medium at a time, we are able to supervise either of $\textbf{M}_c$ or $\textbf{M}_m$ independently.
In this manner, any combination from the learned $\textbf{N}_c$ and $\textbf{N}_m$ are used together to produce the noise for a synthetic image, even from unseen combinations.

We hypothesize that by integrating the noise representation as part of the GOGen, via backpropagation, we should be able to learn latent noise prototypes that are {\it specific to the device but universal across all images captured by that device}. 
Such representations will enable GOGen to better synthesize images under many ($n_c\times n_m$) combinations of sensors and mediums.
We show the learned sensor and medium noise prototypes in Fig.~\ref{fig:sensor_medium_maps}. 
After the input image and noise prototypes are concatenated, they are fed to $8$ convolution layers to synthesize spoof images.
The detailed network architecture of GOGen is shown in Tab.~\ref{table:gogen_golab_structure}.

Since the additional spoof noise should be low-energy, an $\mathcal{L}_2$ loss is employed to minimize the difference between the live image and the image synthesized by the generator. 
This loss helps to limit the magnitude of the noise:
\eqnvspace
\begin{equation}\label{eq:Vq_test}
J_{\text{Vis}} = \|\mathbf{I}-\text{CNN}_{\text{Gen}}(\mathbf{T})\|_2^2.
\end{equation}

\setlength{\tabcolsep}{4pt}
\begin{table}[t!]
\caption{Network architectures of GOGen, GOLab, and GODisc. Resizing is done before concatenation if required. Reshaping is done before the fully connected layers at the end of the GOLab and GODisc networks. All strides are of length $1$. All convolutional kernels are of size $3\!\times\!3$, except for Conv$0$ in Golab and GOPad, which have size $5\!\times\!5$. The dropout rate is $0.5$. For the output, we show the size (height and width) and number of channels.}
\centering
\tiny
\resizebox{\linewidth}{!}{
\begin{tabular}{rllllllll}
\toprule
Method     & \multicolumn{2}{l}{\textbf{GOGen}} && \multicolumn{2}{l}{\textbf{GOLab}} && \multicolumn{2}{l}{\textbf{GODisc}} \\
\cline{2-3} \cline{5-6} \cline{8-9}
Layer      & Inputs        & Output    && Inputs          & Output   && Inputs      & Output \\
\midrule
Img        & -             & $64,3$    && -               & $64,3$   && -           & $64,3$ \\
Lab        & -             & $64,2$    &&                 &          &&             & \\
Conc$0$    & Img,Lab       & $64,5$    &&                 &          &&             & \\
Conv$0$    & Conc$0$       & $64,64$   && Img             & $64,64$  && Img         & $64,32$ \\
Pool$0$    &               &           && Conv$0$         & -        && Conv$0$     & - \\
Conv$1$    & Conv$0$       & $64,96$   && Pool$0$         & $32,96$  && Pool$0$     & $32,32$ \\
Conv$2$    & Conv$1$       & $64,96$   && Conv$1$         & $32,128$ && Conv$1$     & $32,64$ \\
Conv$3$    & Conv$2$       & $64,96$   && Conv$2$         & $32,96$  && Conv$2$     & $32,64$ \\
Pool$1$    &               &           && Conv$3$         & -        && Conv$3$     & - \\
Conv$4$    & Conv$3$       & $64,96$   && Pool$1$         & $16,128$ && Pool$1$     & $16,64$ \\
Conc$1$    & Lab,Conv$0$-$4$& $64,450$ &&                 &          &&             & \\
Conv$5$    & Conc$1$       & $64,160$  && Conv$4$         & $16,156$ && Conv$4$     & $16,96$ \\
Conv$6$    & Conv$5$       & $64,64$   && Conv$5$         & $16,128$ && Conv$5$     & $16,96$ \\
Pool$2$    &               &           && Conv$6$         & -        &&             & \\
Conv$7$    &               &           && Pool$2$         & $8,96$   &&             & \\
Conv$8$    &               &           && Conv$7$         & $8,128$  &&             & \\
Conv$9$    &               &           && Conv$8$         & $8,96$   &&             & \\
Conc$2$    & Lab,Conv$5$-$6$ & $64,226$&& Conv$3$,$6$,$9$ & $8,320$  && Conv$3$,$6$ & $32,160$ \\
Conv$10$   & Conv$2$       & $64,3$    && Conc$2$         & $8,96$   && Conc$2$     & $32,64$\\
Conc$3$    & Img,Conv$10$  & $64,3$    &&                 &          &&             & \\
Conv$11$   &               &           && Conv$10$        & $8,64$   && Conv$10$    & $32,32$ \\
Drop$0$    &               &           && Conv$11$        &          && Conv$11$    & - \\
\hline
           &               &           && \multicolumn{2}{l}{\textbf{Sensor Branch}}     &&             & \\
Conv$12$   &               &           && Drop$0$     & $8,3$            &&             & \\
FC$1$      &               &           && Conv$12$    & $1,512$          && Drop$0$     & $1,256$ \\
FC$2$      &               &           && FC$1$       & $1,7$            && FC$1$       & $1,2$ \\
Soft       &               &           && FC$2$       & $1,7$            && FC$2$       & $1,2$ \\\hline
           &               &           &&  \multicolumn{2}{l}{\textbf{Medium Branch}}   &&         & \\
Conv$13$   &               &           && Drop$0$  & $8,3$               &&      & \\
FC$3$      &               &           && Conv$13$ & $1,512$             &&      & \\
FC$4$      &               &           && FC$3$    & $1,7$               &&      & \\
Soft       &               &           && FC$4$    & $1,7$               &&      & \\
\bottomrule
\end{tabular}
}
\label{table:gogen_golab_structure}
\figvspace
\end{table}
\setlength{\tabcolsep}{1.4pt}

\subsection{GODisc: Discriminator and GAN Losses}
\secvspace

Next, the discriminator GODisc ensures that $\hat{\textbf{I}}$ is visually appealing.
The GODisc network includes $10$ convolution layers and $2$ fully connected layers, shown in Tab.~\ref{table:gogen_golab_structure}.
It outputs the Softmax probability for the two classes, real spoof images vs.~synthesized spoof images.

The training of the GAN follows an alternating training procedure. 
During the training of $\text{CNN}_{Disc}()$, we fix the parameters of $\text{CNN}_{Gen}()$ and use the following loss:
\eqnvspace
\eqnvspace
\begin{multline}\label{eq:Gan_train}
 J_{\text{Disc}_{\text{train}}} = -\mathbb{E}_{\mathbf{I} \in \mathcal{R}} \ \text{log}(\text{CNN}_{\text{Disc}}(\mathbf{I})) \\ -\mathbb{E}_{\mathbf{I} \in \mathcal{L}} \ \text{log}(\|1-\text{CNN}_{\text{Disc}}(\text{CNN}_{\text{Gen}}(\mathbf{T}))\|),
\end{multline}
where $\mathcal{R}$ represents the real spoof images and $\mathcal{L}$ the real live images.
During the training of GOGen, we fix the parameters of $\text{CNN}_{Disc}()$ and use the following loss:
\eqnvspace
\begin{equation}\label{eq:Gen_test}
J_{\text{Disc}_{\text{test}}} = -\mathbb{E}_{\mathbf{I} \in \mathcal{L}} \ \text{log}(\|\text{CNN}_{\text{Disc}}(\text{CNN}_{\text{Gen}}(\mathbf{T}))\|).
\end{equation}

\begin{figure*}[t!]
	\begin{center}
		\includegraphics[width=0.945\textwidth]{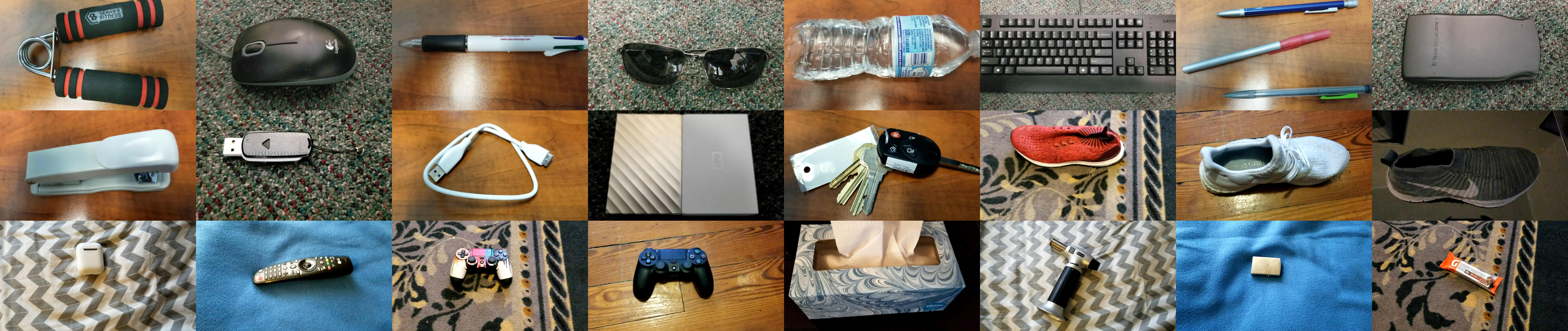}
	\end{center}
	\vspace{-3mm}
	\caption{Example live images of all $24$ objects and $7$ backgrounds from the collected GOSet dataset.}
	\label{fig:all_objects} \figvspace
\end{figure*}

\subsection{GOLab: Sensor and Medium Identification}
\secvspace

GOLab is designed to classify noises from certain sensors and spoof mediums. 
It serves as the discriminator to guide GOGen to generate accurate spoof images as well as the final module to produce scores for GOAS.
Shown in Tab.~\ref{table:gogen_golab_structure}, the input for GOLab is an RGB image with the size of $64\times64$.
The input images can be either the original images or the GOGen synthesized images.
It uses $11$ convolution layers and $3$ max pooling layers to extract features, and then  two fully connected layers to generate $n_c$- and $n_m$-dim vectors for sensor and medium classification.
Each comes from an independent stack of fully connected layers.

We use the cross entropy loss to supervise the training of GOLab.
Given the input image $\textbf{I}$, the ground truth one-hot label $\textbf{a}_m$ and the softmax normalized prediction $\hat{\textbf{a}}_m$ for the spoof medium; and $\textbf{a}_c$, $\hat{\textbf{a}}_c$ for the sensor, the loss functions are defined as:
\eqnvspace
\begin{equation}
S_c=-\sum_i\textbf{a}_c^i\text{log}(\hat{\textbf{a}}_c^i),     \ \quad \   S_m=-\sum_i\textbf{a}_m^i\text{log}(\hat{\textbf{a}}_m^i),
\end{equation}
where $i$ is the class index of the sensors and spoof mediums.
Then, the final loss to supervise GOLab is:
\eqnvspace
\begin{equation}
J_{\text{Lab}_{\text{train}}}=S_c(\textbf{I}) + S_m(\textbf{I}),
\label{eq:golab_train}
\end{equation}

The GOLab network provides supervision for the generator and guides it via backpropagation from the sensor and spoof medium loss functions.
Specifically, we define a normalized loss for updating the generator network:
\eqnvspace
\begin{equation}
J_{\text{Lab}_{\text{test}}}= \frac{S_m(\text{CNN}_{\text{Gen}}(\mathbf{T}))}{1+S_m(\textbf{I})} + \frac{S_c(\text{CNN}_{\text{Gen}}(\mathbf{T}))}{1+S_c(\textbf{I})},\\
\end{equation}
where the numerator shows the classification losses, \emph{i.e.}, $S_m()$ and $S_c()$, for the synthesized images, and $S_m(\textbf{I})$ and $S_c(\textbf{I})$ are the loss of the live images during the updating of GOLab.
By using the normalized loss, GOGen will not be penalized when GOLab has high classification error on the real data, \emph{i.e.}, a large denominator leads to a small quotient regardless of the numerator.

\subsection{GOPad: Binary Classification}
\secvspace

To show the benefits of the proposed method, we follow the baseline algorithm~\cite{liu-siw}, specifically the pseudo-depth map branch, to implement a binary classification of GOAS, termed as GOPad.
To demonstrate strong generalization ability later, we limit the size of the GOPad algorithm by dramatically reducing the number of convolution kernels in each layer to approximately one-third compared to the baseline algorithm.
The GOPad network takes an RGB image as input, and produces a $0$-$1$ map $\text{CNN}_{Pad}(\mathbf{I}) \in \mathbb{R}^{H\times W}$  in the final layer, where it is $0$ for live and $1$ for spoof.
The network activates where the spoof noise is detected.
During the training process, this map allows the CNN model to make live/spoof labeling at the pixel level.
When converged, the $0$-$1$ map should be uniformly $0$ or $1$, representing a confident classification of live vs.~spoof. Formally, the loss function is defined as:
\eqnvspace
\begin{equation}\label{eq:Vpad}
J_{\text{Pad}} = \|\text{CNN}_{\text{Pad}}(\mathbf{I})-\mathbf{G}\|_2^2,
\end{equation}
where $\mathbf{G}$ is the ground truth $0$-$1$ map.

\subsection{Implementation Details} We show all of the three proposed CNN networks in Fig.~\ref{fig:gogen_structure}. We use an alternating training scheme for updating the networks during the training. We train the GOGen while the GOLab and GODisc are fixed. In the next step, we keep the GOGen fixed and train the other two networks. We alternate between these two steps until all networks converge. To train the GOGen and GOLab, we use batch sizes of $40$. Patch sizes of $64\times64$ are used for the GOGen, GODisc, and GOLab.
Patch sizes of $256\times256$ are used for the GOPad, following the setting of previous works.
The final loss for training the generator of GOGen can be summarized as:
\eqnvspace
\begin{equation}
J= J_{\text{Disc}_{\text{test}}} + \lambda_0 J_{\text{Vis}}  + \lambda_1 J_{\text{Lab}_{\text{test}}},\\
\end{equation}
where $\lambda_0$ and $\lambda_1$ are weighting factors. And the final loss for training GODisc and GOLab can be denoted as:
\eqnvspace
\begin{equation}
J= J_{\text{Disc}_{\text{train}}} + \lambda_1 J_{\text{Lab}_{\text{train}}},\\
\end{equation}
and $\lambda_0$ and $\lambda_1$ were set to $0.5$ and $0.1$, for all experiments.

\section{Generic Object Dataset for Anti-Spoofing}
\secvspace

To enable the study of GOAS, we consider a total of $24$ objects, $7$ backgrounds, $7$ commonly used camera sensors, and $7$ spoofing mediums (including live as a blank medium) while collecting the Generic Object Dataset (GOSet). 
If fully enumerated, this would require a prohibitory collection of $8,232$ videos.
Due to constraints, we selectively collect $2,849$ videos to cover most combinations of backgrounds, camera sensors and spoof mediums.

The objects we collect are: squeezer, mouse, multi-pen, sunglasses, water bottle, keyboard, pencils, calculator, stapler, flash drive, cord, hard drive disk, keys, shoe (red), shoe (white), shoe (black), Airpods, remote, PS4 (color), PS4 (black), Kleenex, blow torch, lighter, and energy bar, shown in Fig.~\ref{fig:all_objects}.
Generic objects are more easily available for data collection and are unencumbered by privacy or security concerns, as opposed to human biometrics.
The objects are placed in front of $7$ backgrounds, which are desk wood, carpet speckled, carpet flowered, floor wood, bed sheet (white), blanket (blue), and desk (black).
The spoof mediums include $3$ common computer screens, (Acer Desktop, Dell Desktop, and Acer Laptop), and $3$ mobile device screens, (iPad Pro, Samsung Tab, and Google Pixel), which are of varying size and display quality.

The videos were collected using $7$ commercial devices, (Moto X, Samsung S8, iPad Pro, iPod Touch, Google Pixel, Logitech Webcam, and Canon EOS Rebel).
Except for videos from the iPod Touch at $720$P resolution, all videos are captured at $1,080$P resolution, with average length of $12.5$ seconds.
We first capture the live videos of all objects while varying the distance and viewing angle, and then collect the spoof videos via directly viewing a spoof medium while the live video is displayed on it.
During the collection of spoof videos, care is taken to prevent unnecessary spoofing artifacts (light reflection, screen bezels), as well as data bias (differences in distance, brightness, and orientation).

To leverage the GOSet, we split it into a train and test set.
The train set is composed of the first $13$ objects and corresponds to the first $2$ backgrounds.
The test set is composed of the rest of the objects and backgrounds.
This split prevents overlap and presents a real-world testing scenario.

\begin{table}[t!]
	\caption{Comparison of modality specific anti-spoofing algorithms and GOLab. All methods are trained and tested on GOSet.}
	\centering
	\resizebox{0.8\linewidth}{!}{
	\begin{tabular}{cccc}
		\toprule
		Algorithm                                          & HTER   & EER    & AUC \\
		\midrule
		Chingovska LBP~\cite{replay-attack}                & $16.6$ & $16.9$ & $91.6$ \\
		Boulkenafet Texture~\cite{boulkenafet-ct-using}    & $18.2$ & $19.5$ & $89.1$ \\
		Boulkenafet SURF~\cite{boulkenafet-generalization} & $34.0$ & $35.1$ & $67.6$ \\
		Atoum \emph{et al.}~\cite{atoum}                                 & $13.4$ & $13.5$ & $91.2$ \\
		GOPad (Ours)                                    & $20.6$ & $22.9$ & $87.6$ \\
		GOLab (Ours)                                     & $\textbf{6.3}$  & $\textbf{6.7}$  & $\textbf{97.5}$ \\
		\bottomrule
	\end{tabular}
	}
	\label{table:baseline_performance}
	\figvspace
\end{table}

\section{Experiments}
\secvspace
\label{sec:exp}

In all experiments, we use the training/testing partition mention above to train and evaluate the proposed method.
For evaluation metrics, we report Area Under the Curve (AUC), Half Total Error Rate (HTER)~\cite{hter-paper}, and Equal Error Rate (EER)~\cite{eer-paper}.
Performance is video-based, which is computed via majority voting of patch scores.
For each video, we use all frames; and for each frame, we randomly select $20$ patches.

\begin{table}[t]
  \caption{Confusion matrices for camera sensor and spoof medium identification. The identification accuracy for each sensor/medium and averages are reported using majority voting of $20$ patches from each frame in a video.}
  \resizebox{\linewidth}{!}{
    \begin{tabular}{ccccccccc}
      \toprule
      Sensor  & (1) & (2) & (3) & (4) & (5) & (6)  & (7)  & Acc \\
      \midrule
      (1) Moto X & \cellcolor{blue!12}$16$ & $0$ & $7$ & $5$ & $18$ & $0$ & $0$ & $34.8$ \\
      (2) Logitech & $2$ & \cellcolor{blue!12}$320$ & $0$ & $0$ & $0$ & $0$ & $3$ & $98.5$ \\
      (3) Samsung S8 & $1$ & $2$ & \cellcolor{blue!12}$353$ & $1$ & $0$ & $7$ & $17$ & $92.7$ \\
      (4) iPad Pro & $6$ & $0$ & $42$ & \cellcolor{blue!12}$220$ & $0$ & $3$ & $0$ & $81.2$ \\
      (5) Canon EOS & $55$ & $0$ & $7$ & $32$ & \cellcolor{blue!12}$68$ & $0$ & $3$ & $41.2$ \\
      (6) iPod Touch & $0$ & $0$ & $0$ & $0$ & $0$ & \cellcolor{blue!12}$270$ & $0$ & $100.0$ \\
      (7) Google Pixel & $1$ & $1$ & $0$ & $0$ & $0$ & $1$ & \cellcolor{blue!12}$259$ & $98.9$ \\
      Overall &  &  &  &  &  &  &  & $87.6$ \\\bottomrule
      \multicolumn{9}{c}{(a)}\\
      \vspace{-2mm}
      \\\toprule
      
      Medium  & (1) & (2) & (3) & (4) & (5) & (6)  & (7)  & Acc \\ \midrule
      (1) Live & \cellcolor{blue!12}$97$ & $7$ & $0$ & $0$ & $0$ & $1$ & $0$ & $92.4$ \\
      (2) Acer Desktop & $50$ & \cellcolor{blue!12}$116$ & $67$ & $36$ & $9$ & $45$ & $3$ & $35.6$ \\
      (3) Dell Desktop & $31$ & $52$ & \cellcolor{blue!12}$83$ & $59$ & $20$ & $77$ & $8$ & $25.2$ \\
      (4) Acer Laptop & $58$ & $53$ & $4$ & \cellcolor{blue!12}$141$ & $7$ & $3$ & $5$ & $52.0$ \\
      (5) iPad Pro & $43$ & $30$ & $31$ & $29$ & \cellcolor{blue!12}$107$ & $30$ & $0$ & $39.6$ \\
      (6) Samsung Tab & $4$ & $0$ & $0$ & $79$ & $5$ & \cellcolor{blue!12}$115$ & $0$ & $56.7$ \\
      (7) Google Pixel & $7$ & $54$ & $5$ & $12$ & $34$ & $20$ & \cellcolor{blue!12}$84$ & $38.9$ \\
      Overall &  &  &  &  &  &  &  & $43.2$ \\ \bottomrule
      \multicolumn{9}{c}{(b)}\\
    \end{tabular}
  }
  \figvspace
  \label{table:confusion_matrices}
\end{table}

\begin{figure*}[t!]
  \begin{center}
    \begin{tabular}{@{}c@{}c@{}c@{}c@{}}
      \includegraphics[width=0.25\textwidth]{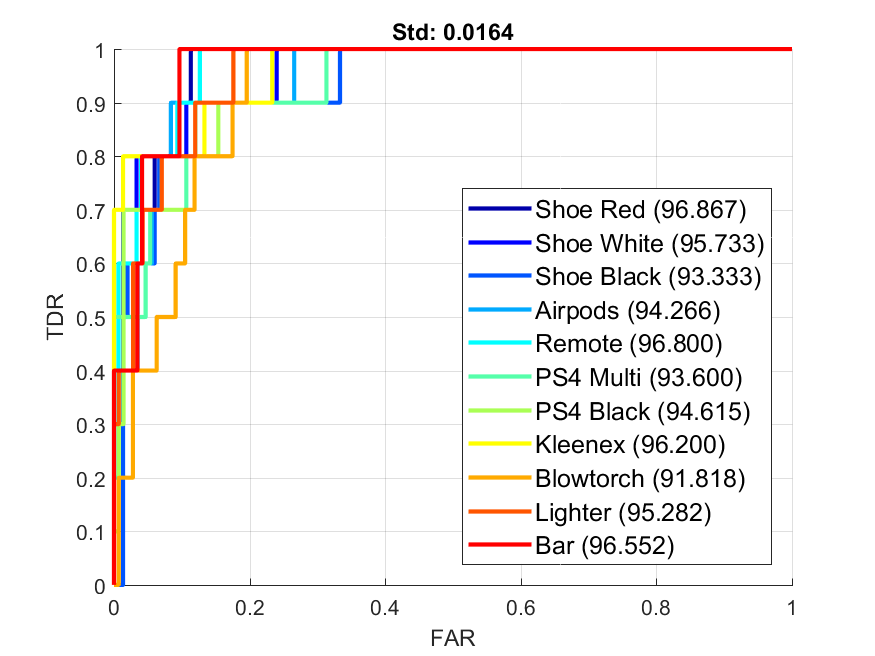} &
      \includegraphics[width=0.25\textwidth]{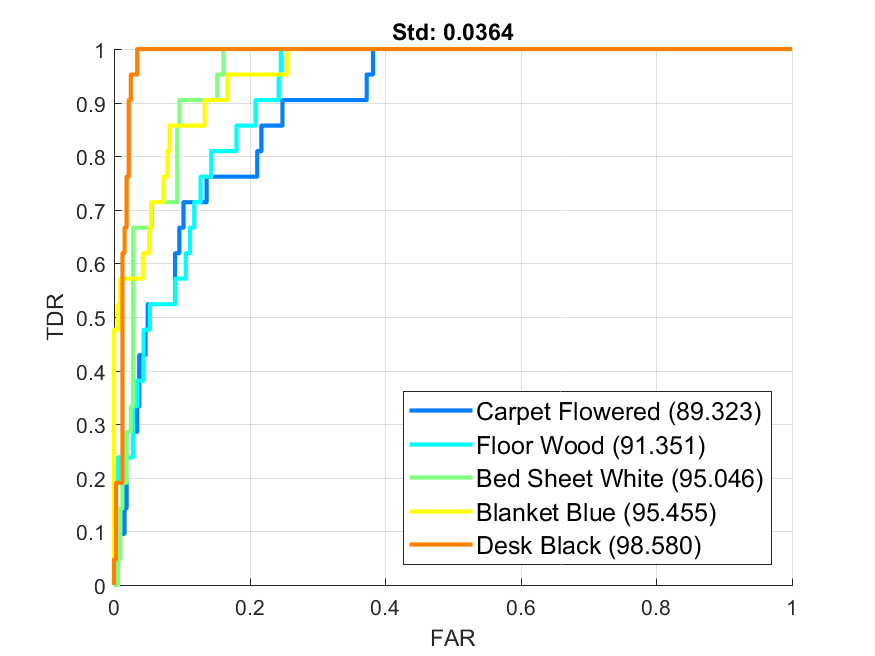} &
      \includegraphics[width=0.25\textwidth]{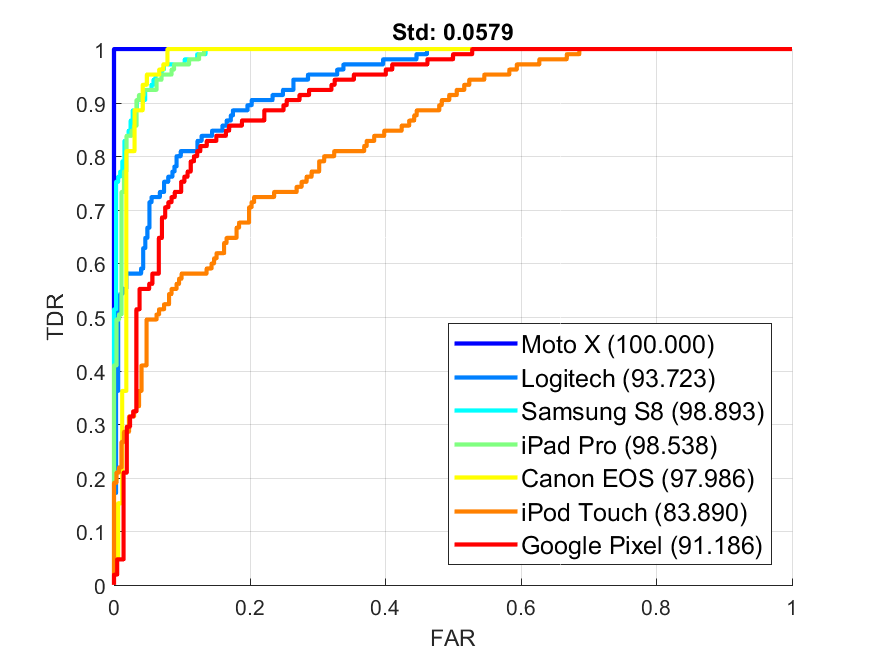} &
      \includegraphics[width=0.25\textwidth]{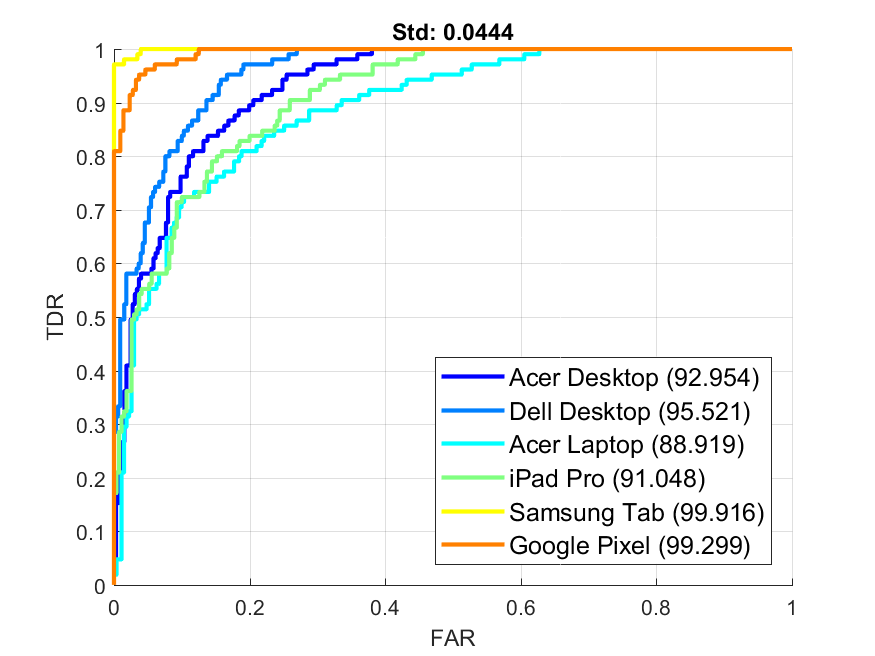} \\ [-1mm]
      \small (a) & (b) & (c) & (d) \\
    \end{tabular}
  \end{center}
  \vspace{-3mm}
  \caption{ROC curves for the anti-spoofing performance of the GOLab algorithm on the GOSet test set. (a) Performance by objects, (b) Performance by backgrounds, (c) Performance by sensors, and (d) Performance by spoof mediums.}
  \label{fig:set_compare}
  \figvspace
\end{figure*}

\subsection{Generic Object Anti-Spoofing}

\textbf{Baseline Performance:}
To demonstrate the superiority of our proposed method, we compare our method with our implementation of the recent methods~\cite{atoum, boulkenafet-ct-using, boulkenafet-generalization, replay-attack} on the GOSet test set.
These recent methods are modality specific algorithms that perform anti-spoofing based on color and texture information.
From Tab.~\ref{table:baseline_performance}, it is shown that GOLab outperforms the other anti-spoofing methods by a large margin for the GOAS task.

\textbf{Benefits of GOLab:}
Tab.~\ref{table:confusion_matrices} (a) and (b) show the confusion matrices of GOLab on sensor and spoof medium classification.
The classification performance for sensors is noticeably better than that of mediums, with the overall accuracy of $87.6\%$ vs.~$43.2\%$.
Although Fig.~\ref{fig:sensor_medium_maps} indicates the noises among medium have distinct patterns, it is worth noting that the medium noises can be ``hidden'' in the image by the sensor noises, which causes the lower accuracy.
The accuracy for detecting live videos is $92.4\%$ which exhibits its promising ability for the anti-spoofing task.

We compute the ROC curves of GoLab on GOSet testing data. Fig.~\ref{fig:set_compare} (a) and (b) show the ROC curves of different objects and different backgrounds respectively.
We can see the AUCs for different objects are similar.
But AUCs for different backgrounds have larger variation, which denotes that the GOLab is more sensitive to surfaces with rich texture, \emph{e.g.}, Carpet Flowered in (b).
By comparing the ROCs for different sensors in Fig.~\ref{fig:set_compare} (c), we observe that the ``Google Pixel'' and ``iPod Touch'' are the hardest sensors to detect, because they are the highest and lowest quality, respectively.
This causes images from the iPod to appear more spoof-like, and images from the Pixel less so, while their respective noise patterns are most distinguishable in Tab.~\ref{table:confusion_matrices}.
Similarly, the ``Acer Laptop'' is the most challenging spoof medium for anti-spoofing, shown in Fig.~\ref{fig:set_compare} (d).

\begin{table}
	\caption{Performance of GOLab when trained on varying amounts of live, real spoof, and synthetic spoof data. Live data was randomly selected. For each live video, $1$ or $2$ (out of $6$ possible) spoof videos were then selected. We randomly select from the generated data to increase the training data by $10\%$.}
	\centering
	\resizebox{0.93\linewidth}{!}{
		\begin{tabular}{ccccccc}
			\toprule
			Data         & \multicolumn{3}{c}{GOLab} & \multicolumn{3}{c}{GOLab + GOGen} \\
			\midrule
			Live, Spoof  & AUC    & HTER   & EER       & AUC    & HTER   & EER \\
			\midrule
			$1/4$, $1/6$ & $79.7$ & $26.8$ & $27.7$    & $79.7$ & $27.2$ & $27.6$ \\
			$1/4$, $1/3$ & $85.1$ & $24.0$ & $25.7$    & $86.5$ & $22.3$ & $22.8$ \\
			$1/2$, $1/6$ & $81.9$ & $24.7$ & $26.7$    & $86.0$ & $22.2$ & $22.8$ \\
			$1/2$, $1/3$ & $87.6$ & $19.6$ & $21.0$    & $92.5$ & $14.9$ & $16.2$ \\
			\bottomrule
		\end{tabular}
	}
	\figvspace
	\label{table:all_roc}
\end{table}

\textbf{ Benefits of GOGen:}
GOGen generates synthetic spoof images and performs data augmentation to improve the training of GOLab. It can synthesize spoof images which may be under-represented or missing in the training data.
To present the advantage of GOGen, we train the GOLab with different compositions of training data.
The data compositions and corresponding results are shown in Tab.~\ref{table:all_roc}.
Comparing the relative performance, we see that more spoof data is more important than more live data because additional spoof data contains sensor and medium noise, whereas live data only has sensor noise.
Comparing the performance of the GOLab without GOGen to those of the GOLab with GOGen, the inclusion of synthetic data during training has significant benefit for the anti-spoofing performance of GOLab.
As additional sensors/mediums are introduced, GOGen can reduce the cost of future data collection by appropriately generating images for the new sensor/medium combinations.

\subsection{Face Anti-Spoofing Performance}
We also evaluate the generalization performance of the proposed method on face anti-spoofing tasks.
We present cross-database testing between two face anti-spoofing databases, SiW and OULU-NPU.
The testing on OULU-NPU follows the Protocol $1$ and the testing on SiW is executed on all test data.
The evaluation and comparison include two parts: firstly, we train the previous methods on either OULU-NPU or SiW, and test on the other; secondly, we train the previous methods and ours on GOSet, and test on the two face databases.
The results are shown in Tab.~\ref{table:gopad_eval_on_face}.
GOPad is structurally very similar to the Atoum~\emph{et al.} algorithm~\cite{atoum}, however, ~\cite{atoum} uses more than $10$X the number of network parameters.
The similar performance between these two methods implies that the leaner and faster GOPad was able to learn strong discriminative ability, regardless of its smaller size.
The SOTA performance of both Atoum~\emph{et al.}  and GOPad on SiW when trained on GOSet demonstrates the generalization ability from generic objects to face data.
The lack of such performance when tested on OULU shows that the generalization of current methods to unseen sensors/mediums is poor, providing future incentive for GOGen to synthesize data that represents these devices.

\begin{table}[t]
  \small
  \begin{center}
    \caption{Performance of GOPad and GOLab algorithms along with SOTA face anti-spoofing algorithms on face anti-spoofing datasets. The algorithms trained on face data are cross-tested between OULU and MSU-SiW. The rest are trained on GOSet. [Key: \textbf{Best}, \textit{Second best}]}
    \label{table:gopad_eval_on_face}
    \resizebox{0.98\linewidth}{!}
    {
      \begin{tabular}{cccccc}
        \toprule
        &       & \multicolumn{2}{c}{OULU P1} & \multicolumn{2}{c}{MSU SiW} \\
        \midrule
        Algorithm                                          & Train & HTER   & EER    & HTER   & EER \\
        \midrule
        Chingovska  LBP~\cite{replay-attack}               & Face  & $38.5$ & $44.2$ & $30.5$ & $31.7$ \\
        Boulkenafet Texture~\cite{boulkenafet-ct-using}    & Face  & $40.8$ & $43.3$ & $28.6$ & $29.9$ \\
        Boulkenafet SURF~\cite{boulkenafet-generalization} & Face  & $38.2$ & $40.8$ & $36.0$ & $36.7$ \\
        Atoum \emph{et al.} ~\cite{atoum}                                 & Face  & $\textbf{11.8}$ & $\textbf{13.3}$ & $11.0$ & $11.2$ \\
        \midrule
        Chingovska LBP~\cite{replay-attack}                & GOSet & $44.1$ & $46.1$ & $42.2$ & $42.4$ \\
        Boulkenafet Texture~\cite{boulkenafet-ct-using}    & GOSet & $34.6$ & $36.7$ & $44.1$ & $44.9$ \\
        Boulkenafet SURF~\cite{boulkenafet-generalization} & GOSet & $45.3$ & $45.8$ & $47.7$ & $48.6$ \\
        Atoum \emph{et al.}~\cite{atoum}                                 & GOSet & $\textit{32.9}$ & $35.0$ & $\textbf{8.2}$  & $\textbf{8.8}$ \\
        GOPad (Ours)                                     & GOSet & $33.4$ & $\textit{34.2}$ & $\textit{9.5}$ & $\textit{10.2}$ \\
        GOLab (Ours)                                     & GOSet & $41.2$ & $42.5$ & $15.6$ & $16.0$ \\
        \bottomrule
      \end{tabular}
    }
  \end{center}
  \figvspace
  \vspace{-3mm}
\end{table}\textsl{}

We train Atoum \emph{et al.}~\cite{atoum} using MSU SiW face dataset and test on the GOSet dataset, resulting in an AUC of $62.3$, HTER of $37.0$, and EER of $41.4$.
Comparing to Tab.~\ref{table:all_roc}, Atoum \emph{et al.}~\cite{atoum} has the lowest performance, even worse than GOLab trained with the smallest amount of data.
This shows that models trained only on faces are domain specific and can not model or detect the true noise in spoof images.

\subsection{Ablation Study}

\textbf{Noise representation:}
Fig.~\ref{fig:sensor_medium_maps} shows the learned noise prototypes for the sensors and mediums. 
In the last row of Fig.~\ref{fig:sensor_medium_maps}, the distinctive high frequency information is evident in the FFT of the spoof medium prototypes.
In contrast, the FFT for the sensor prototypes are similar.
To evaluate the advantage of modeling noise prototypes, we train the GOGen network without noise prototypes by constructing $\textbf{T}=[\textbf{I},\textbf{M}_c^{\prime},\textbf{M}_m^{\prime}]$. 
$\textbf{M}_c^{\prime}$ and $\textbf{M}_m^{\prime}$ are of the same size of as $\textbf{M}_c$ and $\textbf{M}_m$, and with all elements being zeros except the prototypes of selected spoof and medium being all $1$.
The Rank-$1$ accuracy for sensor and spoof medium identification of the related GOLab on the synthesized data is $11.0\%$ and $19.7\%$, respectively.
However, by learning noise prototypes, as shown in Fig.~\ref{fig:gogen_structure}, the accuracy is $56.0\%$ and $26.3\%$.

\begin{table}
  \centering
  \small
  \caption{Anti-spoofing performance of GOPad and GOLab on the GOSet dataset with varying amounts of training data.}
  \resizebox{\linewidth}{!}{
    \begin{tabular}{ccccccc}
      \toprule
      Data              & \multicolumn{3}{c}{Golab} & \multicolumn{3}{c}{GoPad} \\
      \midrule
      (Live, Spoof)     & AUC    & HTER   & EER       & AUC    & HTER   & EER\\
      \midrule
      ($1/4$, $1/6$)    & $79.7$ & $26.8$ & $27.7$     & $84.4$ & $23.8$ & $24.8$ \\
      ($1/4$, $4/6$)    & $86.0$ & $21.6$ & $23.8$    & $86.2$ & $22.4$ & $22.9$ \\
      (All, $4/6$)      & $94.6$ & $12.5$ & $13.9$    & $86.3$ & $22.4$ & $23.8$ \\
      (All, All)        & $97.5$ & $6.3$  & $6.7$    & $87.6$ & $20.6$ & $22.9$ \\\bottomrule
    \end{tabular}
  }
  \label{table:pad_vs_lab}
  \figvspace
\end{table}

\begin{figure}[t!]
	\begin{center}
		\includegraphics[width=\linewidth]{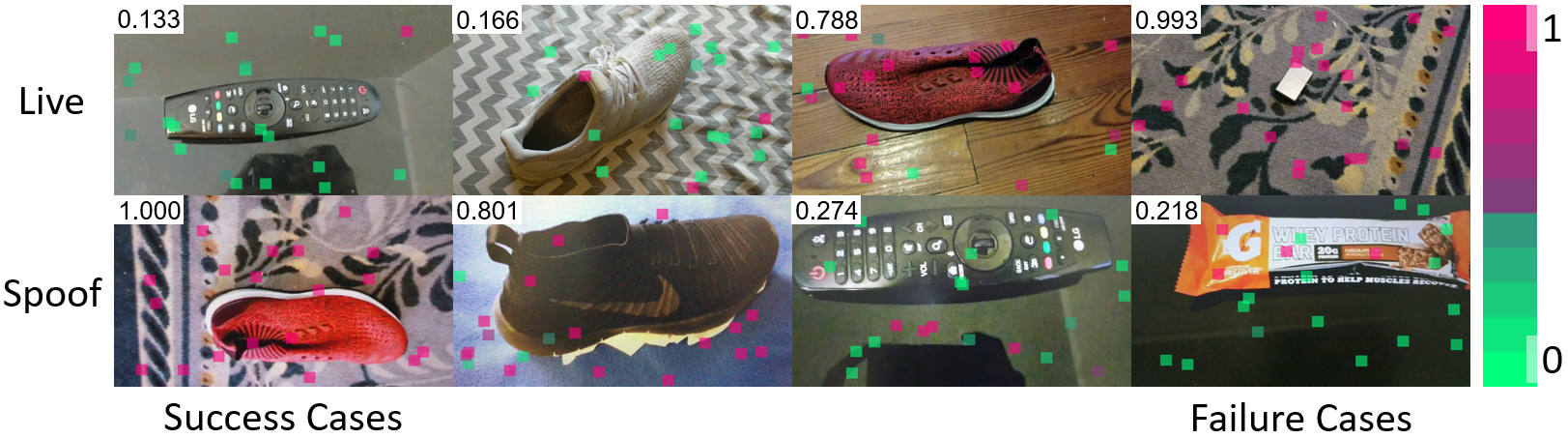}
	\end{center}
	\figvspace
	\caption{Illustration of GOLab-based anti-spoofing, with $2$ success (left) and $2$ failure (right) cases for live (top row) and real spoof (bottom row) using $20$ patches per image. The color bar shows the output range of the network: $1$ is spoof and $0$ is live. The score at the top left corner is the average of all patches.
	}
	\figvspace
	\label{fig:success_failure_cases}
\end{figure}

\textbf{Binary or N-ary Classification:}
We train the GOPad on the GOSet dataset, and we find that GOPad performs better than GOLab when only a small amount of data is utilized for training.
However, GOLab is better than GOPad when using a larger training set.
The training data to be used is chosen by randomly sampling of the GOSet training set.
We attribute this improvement to the auxiliary information (classification between multiple sensors and mediums) that is learned by GOLab for the sensor and spoof medium identification.
The detailed comparison is shown in Tab.~\ref{table:pad_vs_lab}.

\textbf{GOLab Loss Functions:}
To demonstrate the benefit of both sensor and medium classification in the GOLab algorithm, experiments were run using each independently.
Using only $S_c(\textbf{I})$ in Eq.~\ref{eq:golab_train}, we obtain a Rank-$1$ accuracy of $84.7$\%.
Similarly using only $S_m(\textbf{I})$, we obtain an accuracy of $42.0$\% with anti-spoofing performance AUC of $85.9$, HTER of $22.1$ and EER of $22.8$.
By fusing tasks, we improve accuracy for sensor and medium to $87.6$\% and $43.2$\%, respectively.
This also improves anti-spoofing performance to AUC of $97.5$, HTER of $6.3$, and EER of $6.7$.

\subsection{Visualization and Qualitative Analysis}

\begin{figure}[t]
	\begin{center}
		\includegraphics[width=\linewidth]{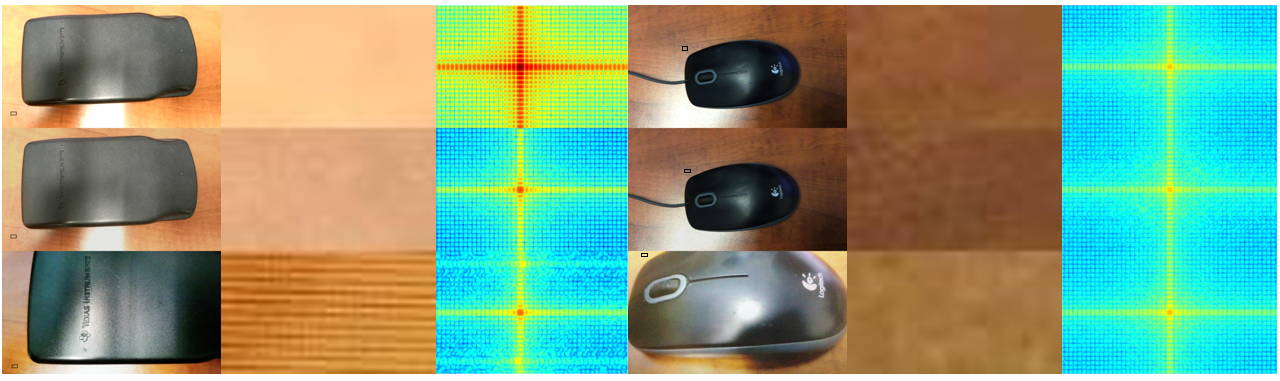}
	\end{center}
	\vspace{-3mm}
	\caption{Visual comparison of live (first row), synthetic spoof (second row), and real spoof (third row) images. Columns are whole image, image patch, and the FFT power spectrum of the image patch. Each synthetic image was generated from a live image. The corresponding ground truth spoof images (third row) are collected with the target sensor/spoof medium combination.}
	\figvspace
	\label{fig:gogen_visualization}
\end{figure}

Fig.~\ref{fig:success_failure_cases} shows success and failure cases of the GOLab model on the GOSet dataset.
This suggests that the smooth, reflective background is classified disproportionately as live and the textured carpet/cloth backgrounds are inversely classified as spoof.
Hence, it is crucial that GOAS and biometric anti-spoofing be possible over the entire image, because no singular patch in the image can provide an accurate and confident score for the entire image.

We show some examples of the generated synthetic spoof images in Fig.~\ref{fig:gogen_visualization}.
We can compare the visual quality with their corresponding live and real spoof images.
The GOGen network is trained to change the high frequency information in the images which are related to the sensor and spoof medium noises.
GOGen is successfully able to alter the high frequency information in these patches to be more similar to the associated spoof than the input live.

\section{Conclusion}
\secvspace

We present our proposed generic object anti-spoofing method which consists of multiple CNNs designed for modeling the sensor and spoof medium noises.
It generates synthetic images which are helpful for increasing anti-spoofing performance.
We show that by modeling the spoof noise properly, the anti-spoofing methods are domain independent and can be utilized in other modalities.
We propose the first generic object anti-spoofing dataset which contains live and spoof videos from $7$ sensors and $7$ spoof mediums.

\paragraph{Acknowledgment}
This research is based upon work supported by the Office of the Director of National Intelligence (ODNI), Intelligence Advanced Research Projects Activity (IARPA), via IARPA R\&D Contract No.~$2017$-$17020200004$. The views and conclusions contained herein are those of the authors and should not be interpreted as necessarily representing the official policies or endorsements, either expressed or implied, of the ODNI, IARPA, or the U.S. Government. The U.S. Government is authorized to reproduce and distribute reprints for Governmental purposes notwithstanding any copyright annotation thereon.

{\small
\bibliographystyle{ieee}
\bibliography{main}

\begin{thebibliography}{10}\itemsep=-1pt

\bibitem{abdelhamed2018high}
A.~Abdelhamed, S.~Lin, and M.~S. Brown.
\newblock A high-quality denoising dataset for smartphone cameras.
\newblock In {\em IEEE Conference on Computer Vision and Pattern Recognition
  (CVPR)}, 2018.

\bibitem{atoum}
Y.~Atoum, Y.~Liu, A.~Jourabloo, and X.~Liu.
\newblock Face anti-spoofing using patch and depth-based {CNN}s.
\newblock In {\em International Joint Conference on Biometrics (IJCB)}, 2017.

\bibitem{bao-liveness-optical-flow}
W.~Bao, H.~Li, N.~Li, and W.~Jiang.
\newblock A liveness detection method for face recognition based on optical
  flow field.
\newblock In {\em International Conference on Image Analysis and Signal
  Processing}, 2009.

\bibitem{hter-paper}
S.~Bengio and J.~Mariéthoz.
\newblock A statistical significance test for person authentication.
\newblock In {\em Odyssey: The Speaker and Language Recognition Workshop},
  2004.

\bibitem{oulu-competition-paper}
Z.~Boulkenafet, J.~Komulainen, Z.~Akhtar, A.~Benlamoudi, D.~Samai, S.~E.
  Bekhouche, A.~Ouafi, F.~Dornaika, A.~Taleb-Ahmed, L.~Qin, F.~Peng, L.~B.
  Zhang, M.~Long, S.~Bhilare, V.~Kanhangad, A.~Costa-Pazo,
  E.~Vazquez-Fernandez, D.~Perez-Cabo, J.~J. Moreira-Perez,
  D.~Gonzalez-Jimenez, A.~Mohammadi, S.~Bhattacharjee, S.~Marcel, S.~Volkova,
  Y.~Tang, N.~Abe, L.~Li, X.~Feng, Z.~Xia, X.~Jiang, S.~Liu, R.~Shao, P.~C.
  Yuen, W.~R. Almeida, F.~Andalo, R.~Padilha, G.~Bertocco, W.~Dias, J.~Wainer,
  R.~Torres, A.~Rocha, M.~A. Angeloni, G.~Folego, A.~Godoy, and A.~Hadid.
\newblock A competition on generalized software-based face presentation attack
  detection in mobile scenarios.
\newblock In {\em IEEE International Joint Conference on Biometrics (IJCB)},
  2017.

\bibitem{boulkenafet2015face}
Z.~Boulkenafet, J.~Komulainen, and A.~Hadid.
\newblock Face anti-spoofing based on color texture analysis.
\newblock In {\em IEEE International Conference on Image Processing (ICIP)},
  2015.

\bibitem{boulkenafet-ct-using}
Z.~Boulkenafet, J.~Komulainen, and A.~Hadid.
\newblock Face spoofing detection using colour texture analysis.
\newblock {\em IEEE Transactions on Information Forensics and Security (TIFS)},
  2016.

\bibitem{boulkenafet-generalization}
Z.~Boulkenafet, J.~Komulainen, and A.~Hadid.
\newblock On the generalization of color texture-based face anti-spoofing.
\newblock {\em Image Vision Compututing}, 2018.

\bibitem{chen-ensemble}
C.~Chen and M.~C. Stamm.
\newblock Camera model identification framework using an ensemble of
  demosaicing features.
\newblock In {\em IEEE International Workshop on Information Forensics and
  Security (WIFS)}, 2015.

\bibitem{chen2020}
C.~Chen, Z.~Xiong, X.~Liu, and F.~Wu.
\newblock Camera trace erasing.
\newblock In {\em IEEE Computer Vision and Pattern Recognition (CVPR)}, 2020.

\bibitem{fsrnet-end-to-end-learning-face-super-resolution-with-facial-priors}
Y.~Chen, Y.~Tai, X.~Liu, C.~Shen, and J.~Yang.
\newblock {FSRNet}: End-to-end learning face super-resolution with facial
  priors.
\newblock In {\em IEEE Computer Vision and Pattern Recognition (CVPR)}, 2018.

\bibitem{chetty-audio-visual}
G.~Chetty and M.~Wagner.
\newblock Audio-visual multimodal fusion for biometric person authentication
  and liveness verification.
\newblock In {\em NICTA-HCSNet Multimodal User Interaction Workshop}, 2006.

\bibitem{replay-attack}
I.~Chingovska, A.~Anjos, and S.~Marcel.
\newblock On the effectiveness of local binary patterns in face anti-spoofing.
\newblock In {\em International Conference of Biometrics Special Interest Group
  (BIOSIG)}, 2012.

\bibitem{star-gan}
Y.~Choi, M.~Choi, M.~Kim, J.~Ha, S.~Kim, and J.~Choo.
\newblock Star{GAN}: Unified generative adversarial networks for multi-domain
  image-to-image translation.
\newblock In {\em IEEE Conference on Computer Vision and Pattern Recognition
  (CVPR)}, 2017.

\bibitem{chugh2018fingerprint}
T.~Chugh, K.~Cao, and A.~K. Jain.
\newblock Fingerprint spoof buster: Use of minutiae-centered patches.
\newblock {\em IEEE Transactions on Information Forensics and Security (TIFS)},
  2018.

\bibitem{pinto-visual-rhythm}
A.~d.~S.~Pinto, H.~Pedrini, W.~Schwartz, and A.~Rocha.
\newblock Video-based face spoofing detection through visual rhythm analysis.
\newblock In {\em Conference on Graphics, Patterns and Images (SIBGRAPI)},
  2012.

\bibitem{dong-super-res}
C.~Dong, C.~C. Loy, K.~He, and X.~Tang.
\newblock Learning a deep convolutional network for image super-resolution.
\newblock In {\em European Conference on Computer Vision (ECCV)}, 2014.

\bibitem{feng-motion-cues}
L.~Feng, L.~Po, Y.~Li, X.~Xu, F.~Yuan, T.~C.-H. Cheung, and K.~Cheung.
\newblock Integration of image quality and motion cues for face anti-spoofing:
  A neural network approach.
\newblock {\em Journal of Visual Communication and Image Representation}, 2016.

\bibitem{filler-icip}
T.~Filler, J.~Fridrich, and M.~Goljan.
\newblock Using sensor pattern noise for camera model identification.
\newblock In {\em IEEE International Conference on Image Processing (ICIP)},
  2008.

\bibitem{gloe2010dresden}
T.~Gloe and R.~B{\"o}hme.
\newblock The {D}resden image database for benchmarking digital image
  forensics.
\newblock In {\em ACM Symposium on Applied Computing}, 2010.

\bibitem{miroslav-spie}
M.~Goljan, J.~Fridrich, and T.~Filler.
\newblock Large scale test of sensor fingerprint camera identification.
\newblock In {\em SPIE Conference on Electronic Imaging, Security and Forensics
  of Multimedia Contents}, 2009.

\bibitem{guera-counter-forensic}
D.~G{\"{u}}era, Y.~Wang, L.~Bondi, P.~Bestagini, S.~Tubaro, and E.~J. Delp.
\newblock A counter-forensic method for {CNN}-based camera model
  identification.
\newblock In {\em IEEE conference on computer vision and pattern recognition
  workshops (CVPRW)}, 2017.

\bibitem{jourabloo}
A.~Jourabloo, Y.~Liu, and X.~Liu.
\newblock Face de-spoofing: Anti-spoofing via noise modeling.
\newblock In {\em European Conference on Computer Vision (ECCV)}, 2018.

\bibitem{lai-image-sr}
W.~Lai, J.~Huang, N.~Ahuja, and M.~Yang.
\newblock Fast and accurate image super-resolution with deep {L}aplacian
  pyramid networks.
\newblock {\em IEEE Transactions on Pattern Analysis and Machine Intelligence
  (PAMI)}, 2018.

\bibitem{3dmask-rppg}
S.~Liu, P.~C. Yuen, S.~Zhang, and G.~Zhao.
\newblock 3{D} mask face anti-spoofing with remote photoplethysmography.
\newblock In {\em European Conference on Computer Vision (ECCV)}, 2016.

\bibitem{liu-siw}
Y.~Liu, A.~Jourabloo, and X.~Liu.
\newblock Learning deep models for face anti-spoofing: Binary or auxiliary
  supervision.
\newblock In {\em IEEE Computer Vision and Pattern Recognition (CVPR)}, 2018.

\bibitem{Liu_2019_CVPR}
Y.~Liu, J.~Stehouwer, A.~Jourabloo, and X.~Liu.
\newblock Deep tree learning for zero-shot face anti-spoofing.
\newblock In {\em IEEE Conference on Computer Vision and Pattern Recognition
  (CVPR)}, 2019.

\bibitem{Lucena2017TransferLU}
O.~Lucena, A.~Junior, V.~Moia, R.~Souza, E.~Valle, and R.~de~Alencar~Lotufo.
\newblock Transfer learning using convolutional neural networks for face
  anti-spoofing.
\newblock In {\em International Conference Image Analysis and Recognition},
  2017.

\bibitem{conditional-gan}
M.~Mirza and S.~Osindero.
\newblock Conditional generative adversarial nets.
\newblock {\em CoRR}, 2014.

\bibitem{pan-eyeblink}
G.~Pan, L.~Sun, Z.~Wu, and S.~Lao.
\newblock Eyeblink-based anti-spoofing in face recognition from a generic
  webcamera.
\newblock In {\em IEEE International Conference on Computer Vision (ICCV)},
  2007.

\bibitem{patel-face-unlock}
K.~Patel, H.~Han, and A.~K. Jain.
\newblock Secure face unlock: Spoof detection on smartphones.
\newblock {\em IEEE Transactions on Information Forensics and Security (TIFS)},
  2016.

\bibitem{patel-moire-pattern}
K.~Patel, H.~Han, A.~K. Jain, and G.~Ott.
\newblock Live face video vs. spoof face video: Use of moiré patterns to
  detect replay video attacks.
\newblock In {\em International Conference on Biometrics (ICB)}, 2015.

\bibitem{Perez-Cabo_2019}
D.~Perez-Cabo, D.~Jimenez-Cabello, A.~Costa-Pazo, and R.~J. Lopez-Sastre.
\newblock Deep anomaly detection for generalized face anti-spoofing.
\newblock In {\em IEEE Conference on Computer Vision and Pattern Recognition
  (CVPR)}, 2019.

\bibitem{reddy-pulse-oxiometry}
P.~V. Reddy, A.~Kumar, S.~M.~K. Rahman, and T.~S. Mundra.
\newblock A new method for fingerprint anti-spoofing using pulse oxiometry.
\newblock In {\em IEEE International Conference on Biometrics: Theory,
  Applications, and Systems (BTAS)}, 2007.

\bibitem{song-image-sr}
Q.~Song, R.~Xiong, D.~Liu, Z.~Xiong, F.~Wu, and W.~Gao.
\newblock Fast image super-resolution via local adaptive gradient field
  sharpening transform.
\newblock {\em IEEE Transactions on Image Processing}, 2018.

\bibitem{image-super-resolution-via-deep-recursive-residual-network}
Y.~Tai, J.~Yang, and X.~Liu.
\newblock Image super-resolution via deep recursive residual network.
\newblock In {\em IEEE Computer Vision and Pattern Recognition (CVPR)}, 2017.

\bibitem{memnet-a-persistent-memory-network-for-image-restoration}
Y.~Tai, J.~Yang, X.~Liu, and C.~Xu.
\newblock {MemNet}: A persistent memory network for image restoration.
\newblock In {\em International Conference on Computer Vision (ICCV)}, 2017.

\bibitem{thai-heteroscedastic}
T.~H. Thai, R.~Cogranne, and F.~Retraint.
\newblock Camera model identification based on the heteroscedastic noise model.
\newblock {\em IEEE Transactions on Image Processing}, 2014.

\bibitem{thai-cmi}
T.~H. Thai, F.~Retraint, and R.~Cogranne.
\newblock Camera model identification based on the generalized noise model in
  natural images.
\newblock {\em Digital Signal Processing}, 2016.

\bibitem{disentangled-representation-learning-gan-for-pose-invariant-face-recognition}
L.~Tran, X.~Yin, and X.~Liu.
\newblock Disentangled representation learning {GAN} for pose-invariant face
  recognition.
\newblock In {\em Proceeding of IEEE Computer Vision and Pattern Recognition},
  2017.

\bibitem{eer-paper}
K.~P. Tripathi.
\newblock A comparative study of biometric technologies with reference to human
  interface.
\newblock {\em International Journal of Computer Applications}, 2011.

\bibitem{Yang_2019_CVPR}
X.~Yang, W.~Luo, L.~Bao, Y.~Gao, D.~Gong, S.~Zheng, Z.~Li, and W.~Liu.
\newblock Face anti-spoofing: Model matters, so does data.
\newblock In {\em IEEE Conference on Computer Vision and Pattern Recognition
  (CVPR)}, 2019.

\bibitem{feature-transfer-learning-for-face-recognition-with-under-represented-data}
X.~Yin, X.~Yu, K.~Sohn, X.~Liu, and M.~Chandraker.
\newblock Feature transfer learning for face recognition with under-represented
  data.
\newblock In {\em IEEE Computer Vision and Pattern Recognition (CVPR)}, 2019.

\bibitem{cycle-gan}
J.~Zhu, T.~Park, P.~Isola, and A.~A. Efros.
\newblock Unpaired image-to-image translation using cycle-consistent
  adversarial networks.
\newblock In {\em IEEE International Conference on Computer Vision (ICCV)},
  2017.

\end{thebibliography}
}

\end{document}